\documentclass[english,letter]{ieeje}

\usepackage{graphicx}
\usepackage{latexsym}
\usepackage[fleqn]{amsmath}
\usepackage[varg]{txfonts}

\usepackage{cite}
\usepackage{enumerate}
\usepackage{url}
\usepackage{multirow}
\usepackage{CJKutf8} 

\usepackage{comment}
\usepackage{color}

\usepackage{bm}

\FIELD{}
\YEAR{2011}
\NO{1}
\title{Neural Radiance Field Image Refinement through End-to-End Sampling Point Optimization}
\authorlist{%
 \authorentry{Kazuhiro Ohta}{n}{labelA}
 \authorentry[ono@ibe.kagoshima-u.ac.jp]{Satoshi Ono}{n}{labelA}
}
\affiliate[labelA]{%
Information Science and Biomedical Engineering Program, 
Kagoshima University\\
Korimoto 1-21-40, Kagoshima, 890-0065 Japan
}

\newcommand{\ohta}[1]{}
\newcommand{\del}[1]{}

\newcommand{\jtextd}[1]{} 
\newcommand{\jtext}[1]{}
\newcommand{\jtextdmori}[1]{} 
\newcommand{\jtextdmae}[1]{} 

\begin{document}
\begin{abstract}
Neural Radiance Field (NeRF), capable of synthesizing high-quality
novel viewpoint images, suffers from issues like artifact occurrence due
to its fixed sampling points during rendering.
This study proposes a method that optimizes sampling points to
reduce artifacts and produce more detailed images.
\end{abstract}
\maketitle


\section{Introduction}

Recent advancements in neural networks led to increased attention
toward neural radiance fields (NeRF), a technique designed to capture
the color and density of objects in 3D
space~\cite{mildenhall2021nerf}.
%
The use of NeRF facilitates the seamless representation of
3D scenes by estimating color and density of multiple points, thereby
enabling the generation of high-quality images from novel viewpoints.

Rendering using NeRF involves estimating the color information and
probability density at sample points along each ray, thereby
estimating the expected color value for each pixel.
A key consideration here is that the positions of these sampling
points influence the quality of the images produced.
Canonical NeRF places these points at regular intervals,
regardless of the scene's attributes, resulting in artifacts near
thin or light objects.

This study proposes a method for optimizing sampling points along a
ray tailored to the characteristics of the target scene.
The proposed method leverages an architecture inspired by
MLP-Mixer~\cite{tolstikhin2021mlp} to dynamically configure sampling
points within NeRF, capturing scene surfaces and mitigating
artifacts.
%
The direct integration of MLP-Mixer like architecture into NeRF for
sampling point arrangement enables seamless end-to-end learning,
similar to conventional NeRF methodologies.
Experiments using real image datasets indicate that this method
successfully reduces artifacts and enhances image quality relative to
the conventional NeRF method.


\section{Related work}

%
Barron et al. proposed Mip-NeRF,
which represents camera rays as cones instead of lines and computes
color and density within a frustum to reduce artifacts~\cite{barron2021mip}.
%
%
%
Kurz et al. proposed AdaNeRF, which aims to speed up rendering by
reducing the number of sampling points~\cite{kurz2022adanerf}.
As represented by the methods discussed above,
extensive research has been conducted on 
reducing NeRF artifacts and
enhancing rendering speed;
however, the effectiveness of improving rendered image quality through
a dense concentration of sampling points around the objects, as in the
proposed method, has yet to be adequately investigated.

\begin{figure}[t]
  \centering
      \includegraphics[width=.9\linewidth]{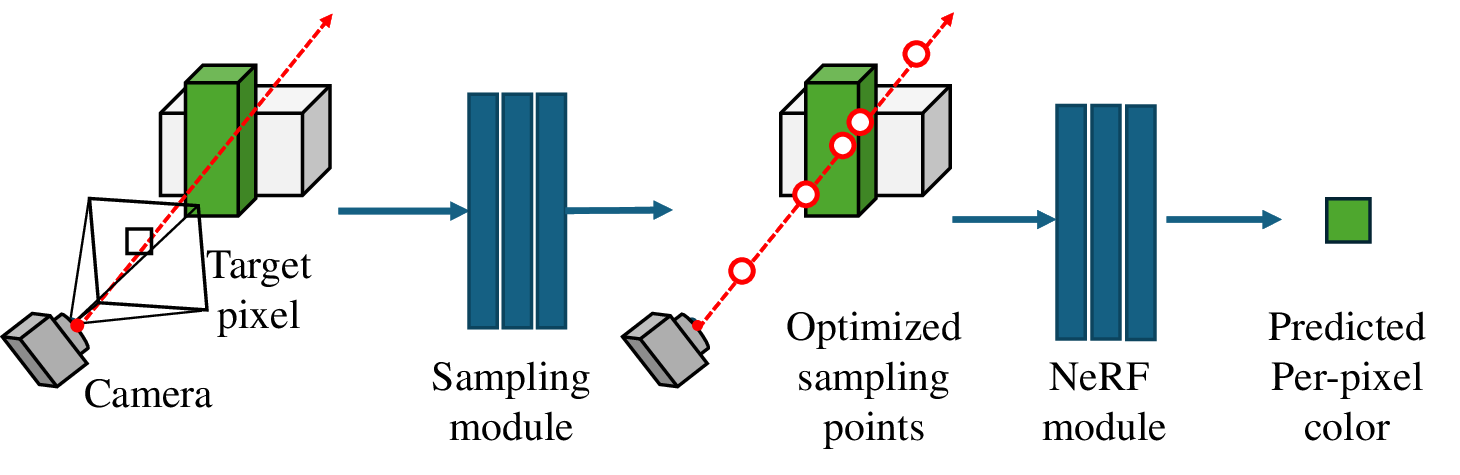}
      \caption{Structure of the proposed method.}
      \label{fig:network_ark}
   ~\\ ~\\
  \centering
    \centering
    \includegraphics[scale=0.26]{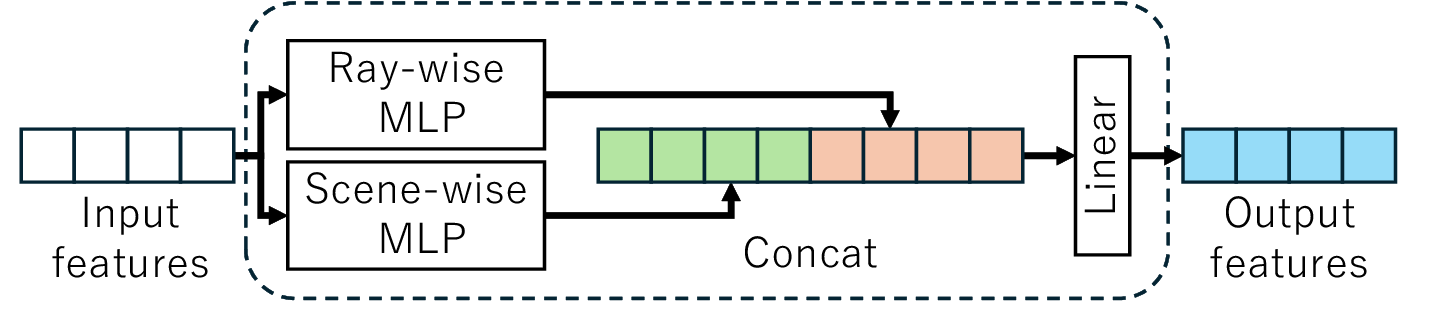}
    \caption{Network structure of the sampling block.}
    \label{fig:sampling}
    ~\\ ~\\
    \begin{tabular}{@{}c@{~~~~~}c@{}}
    \includegraphics[scale=0.24]{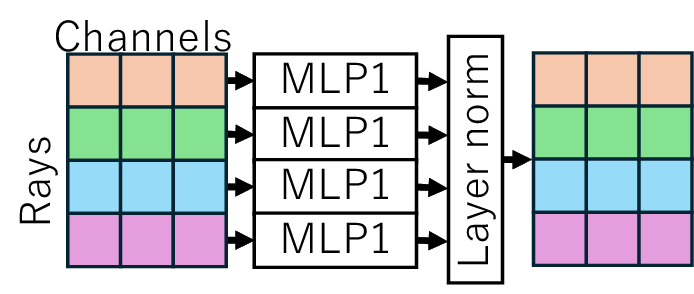} &
    \includegraphics[scale=0.24]{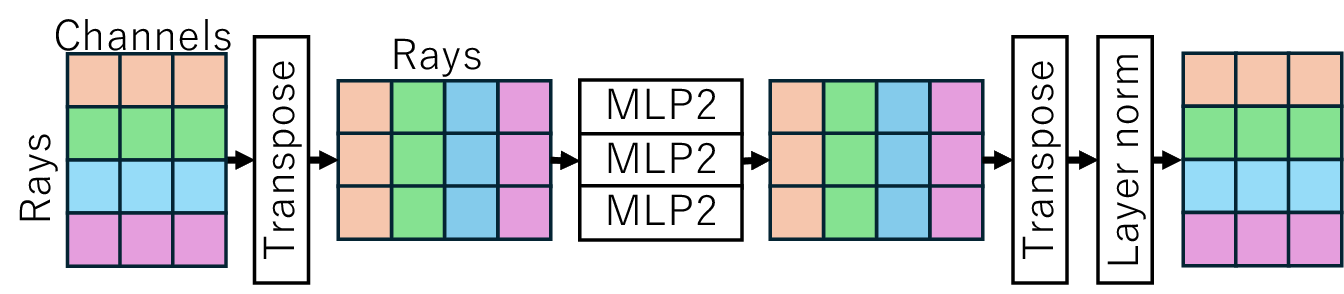}\\
    {\scriptsize (a) Ray-wise MLP} &
    {\scriptsize (b) Scene-wise MLP} \\
    \end{tabular}
    \caption{Architectures of the ray- and scene-wise MLPs.}
    \label{fig:mlp}
\end{figure}

\begin{figure*}[t]
  \centering
      \begin{tabular}{@{}c@{~}c@{~}c@{~}c@{}}
      {\includegraphics[width=0.25\textwidth]{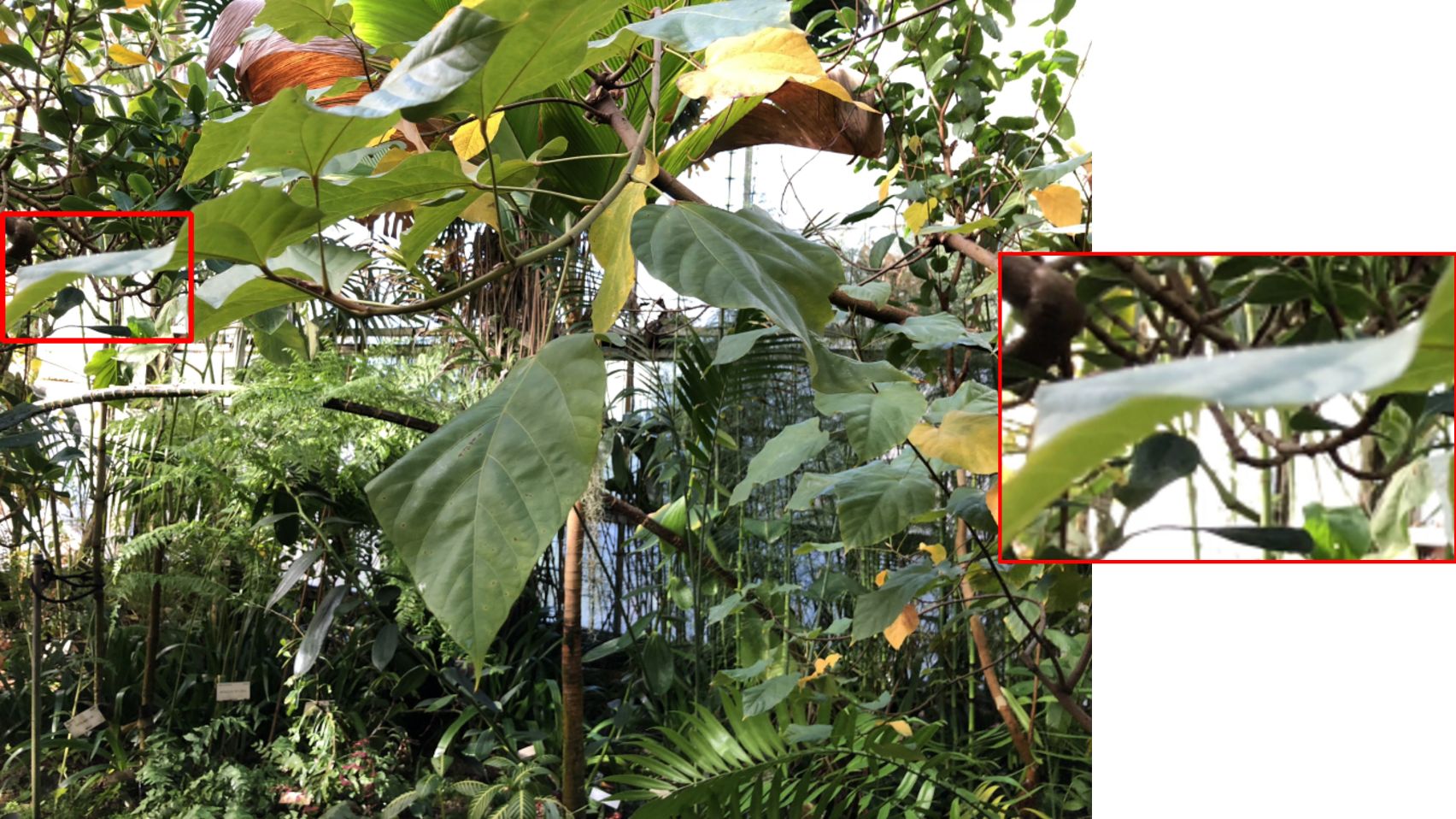}} &
      {\includegraphics[width=0.25\textwidth]{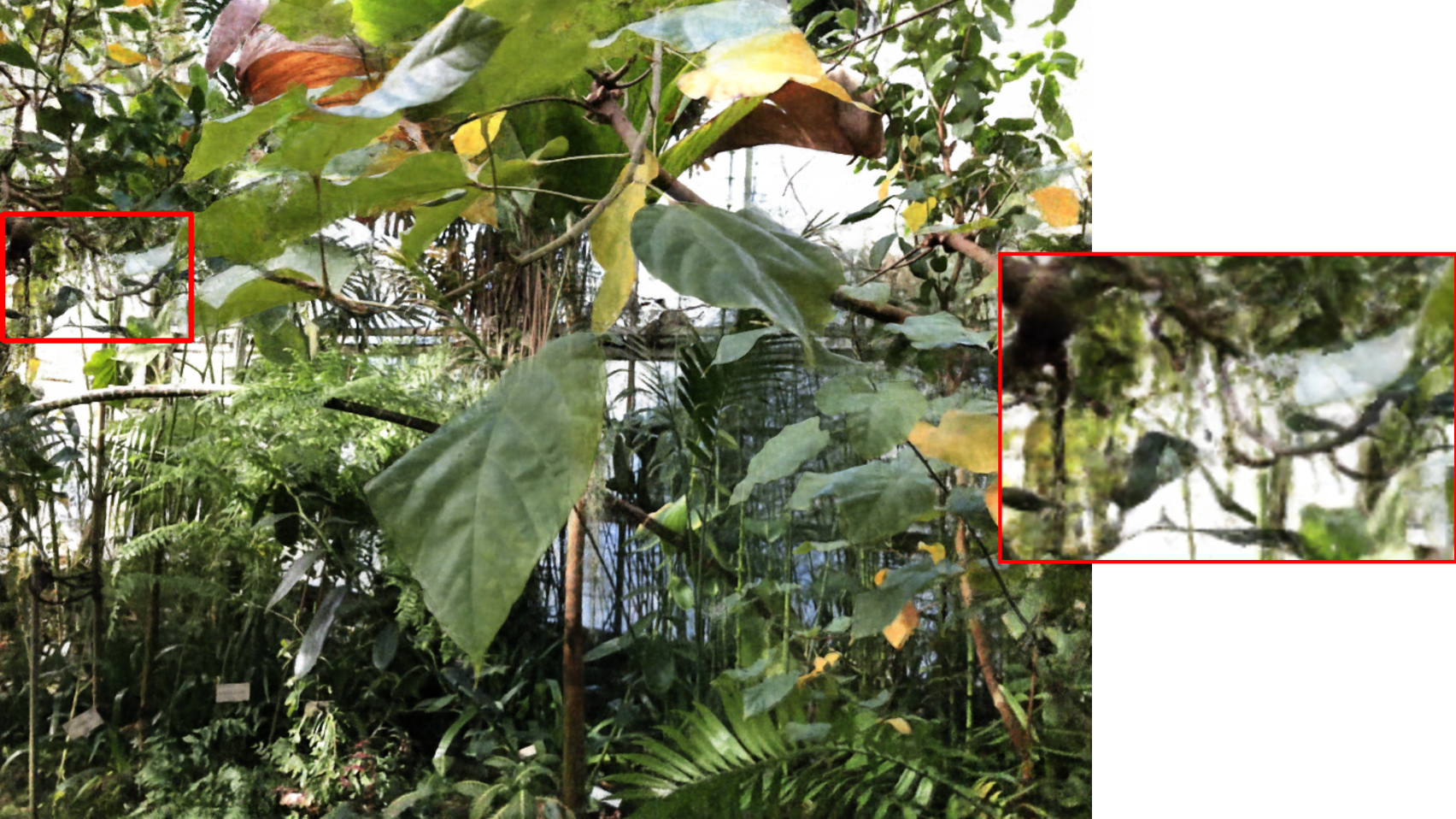}} &
      {\includegraphics[width=0.25\textwidth]{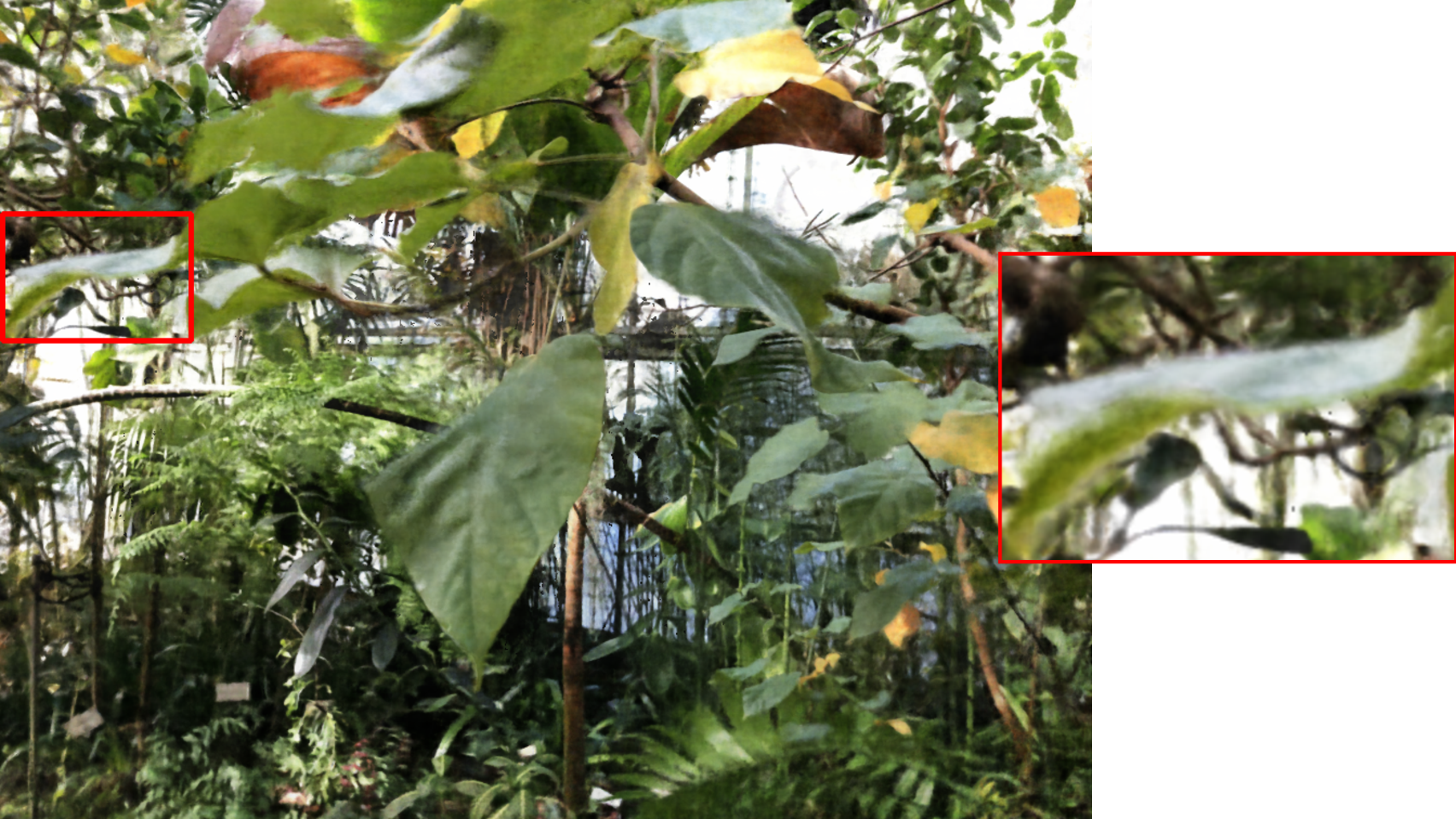}} & 
      {\includegraphics[keepaspectratio, width=0.24\linewidth]{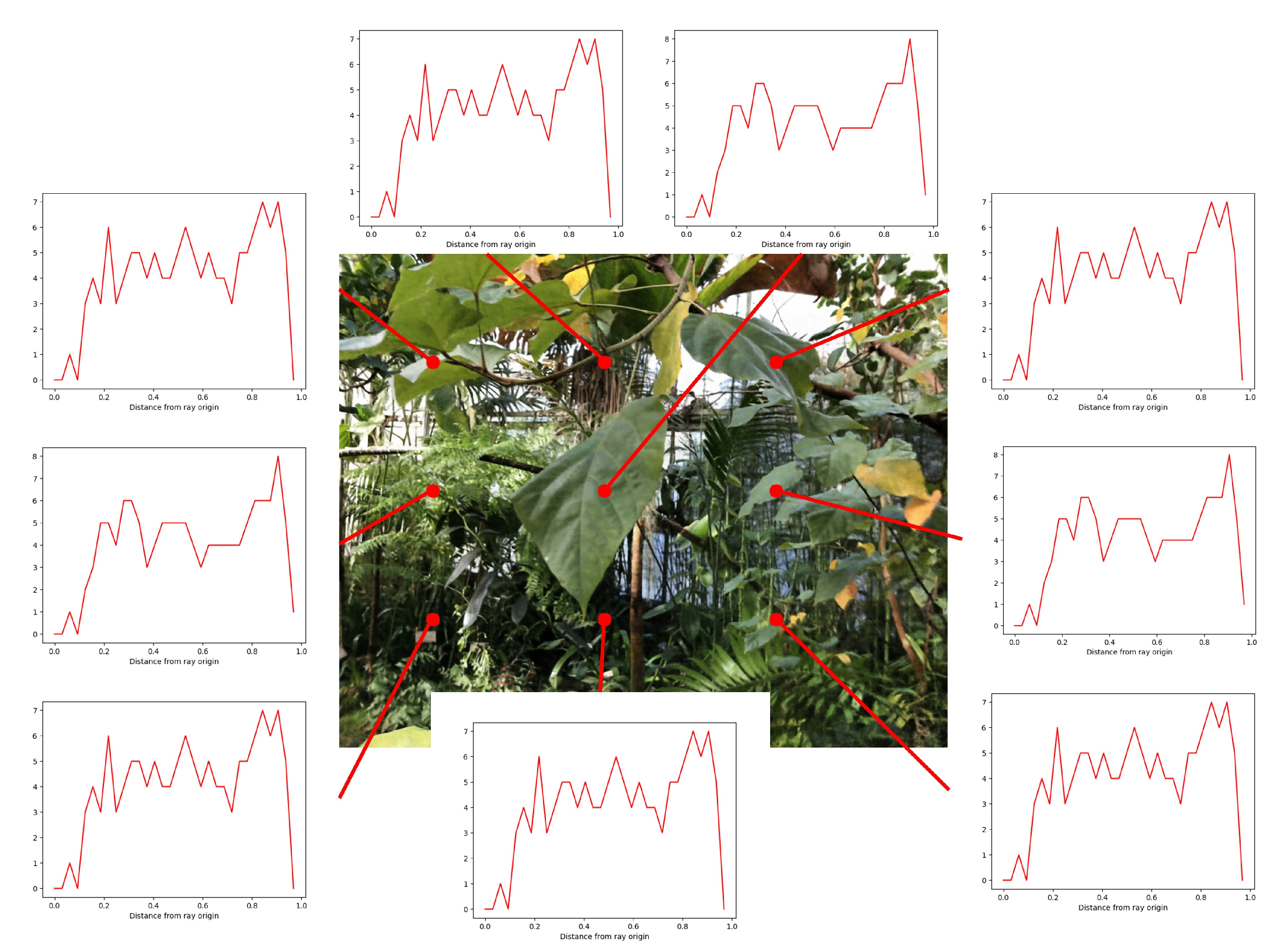}}  \\

       (a) Ground-Truth & (b) Rendering result (NeRF) & (c) Rendering result (Ours) &
       (d) Distributions of sampling points \\
      \end{tabular}
      \caption{Examples of rendered images for test view on Real Foward-Facing dataset.}
      \label{fig:result2}
\end{figure*}


\section{The proposed method}
\subsection{Key idea}

This study proposes a novel viewpoint image generation method using
NeRF with optimized 
sampling points.
The proposed method incorporates a 
MLP-Mixer-based architecture
into NeRF module in a
cascading manner, allowing simultaneous selection of optimal sampling
points and NeRF training in an end-to-end manner.
We anticipate the reduction of artifacts during rendering by
concentrating sampling points primarily in critical areas,
particularly object surfaces.

\subsection{Overview of the proposed method}

Fig.~\ref{fig:network_ark} illustrates the structure of the proposed
method, which
%
employs a cascaded design consisting of a sampling module
and a NeRF module.
The sampling module takes as input the ray information, composed of
camera center
$\bm{o}$ 
and ray direction vector
$\bm{d}$
passing through each pixel of a rendered image,
to estimate appropriate sampling points on the ray.
%

The sampling module's output serves as input for a conventional NeRF
network.
This NeRF module processes these inputs, including the estimated
sampling points and direction vectors of camera rays as in the
conventional NeRF, to estimate the color and density at those points.
These calculated values are then used for volume rendering, leading to
determining individual pixel colors and generating an image based on
the given camera pose.

The model's training process adheres to the conventional NeRF
methodology.
%

\subsection{Structure of the sampling module}

The sampling module $G$ of the proposed method takes as input
the camera center $\bm{o}$ and the direction vectors $\bm{d}_j$ (for $j
\in \{1, \ldots, N_r\}$) for $N_r$ rays in the scene, and outputs the
distances $t_{j,i}$ (for $i \in \{1, \ldots, N_s \}$) from the origin
to the sampling points $i$ on the rays $j$.
The sampling module $G$ is structured with multiple layers of
sampling blocks shown in Fig.~\ref{fig:sampling}, consisting of two
integrated streams: a ray-wise MLP $G^{(ray)}$ for computing
information along the camera rays and a scene-wise MLP
$G^{(scene)}$ for calculating the information between the
camera rays within the scene.

Ray-wise MLP $G^{(ray)}$ operates as a network that processes
the features inherent to individual rays.
%
Fig.~\ref{fig:mlp}(a) illustrates the architecture of
$G^{(ray)}$.
%
The MLP1 network comprises two fully connected layers, with the first
layer's output subjected to GELU activation function.

Scene-wise MLP $G^{(scene)}$ serves as a network that computes
features based on the relationships among all rays within the scene,
adopting a mechanism that allows consideration of the interrelations
among feature vectors possessed by multiple rays.
Fig.~\ref{fig:mlp}(b) shows the architecture of
$G^{(scene)}$.
The structure of MLP2 mirrors that of MLP1.

%
The features processed by the two MLPs are concatenated and
subsequently refined by a linear layer to reduce them to the same
dimension as the input.
Finally, the sampling module yields distances of the sampling points
$\{t_{j,i}\}$, converts them into 3D points $\bm{x}_{j,i}$, and then
enters them into the NeRF module together with ray information
$\bm{d}_j$.

\begin{table}[t]
  \caption{Results on Real Forward-Facing dataset.}
  \label{table:result}
  \centering
  \begin{tabular}{@{~}l@{~}|@{~}l@{~}|@{~}c@{~}c@{~}c@{~}c@{~}c@{~}c@{~}c@{~}c@{~}|@{~}c@{~}}
    \hline
    & & Room & Fern & Leaves & Fortress & Orchids & Flower & T-Rex & Horns & avg \\
    \hline
    PSNR
    & NeRF & $31.79$ & $24.99$ & $20.83$ & $31.06$ & $\bm{20.39}$ & $27.43$ & $26.19$ & $26.97$ & $26.82$ \\
    & Ours & $\bm{32.29}$ & $\bm{25.10}$ & $\bm{20.96}$ & $31.06$ & $19.98$ & $27.44$ & $\bm{26.95}$ & $27.08$ & $\bm{27.01}$ \\
    \hline
    SSIM 
    &NeRF & $0.938$ & $0.780$ & $0.693$ & $0.870$ & $\bm{0.639}$ & $0.824$ & $0.867$ & $0.806$ & $0.819$\\
    &Ours & $\bm{0.943}$ & $\bm{0.790}$ & $\bm{0.707}$ & $0.878$ & $0.623$ & $\bm{0.827}$ & $\bm{0.884}$ & $\bm{0.813}$ & $\bm{0.824}$\\
    \hline
    LPIPS
    &NeRF & $0.124$ & $0.255$ & $0.293$ & $\bm{0.142}$ & $\bm{0.302}$ & $0.198$ & $0.187$ & $0.257$ & $0.212$ \\
    &Ours & $\bm{0.111}$ & $\bm{0.243}$ & $\bm{0.278}$ & $0.153$ & $0.312$ & $0.196$ & $\bm{0.168}$ & $\bm{0.243}$ & $\bm{0.204}$ \\
    \hline
  \end{tabular}
\end{table}

\section{Evaluation}

Experiments were conducted on the real Forward-Facing dataset
\cite{mildenhall2021nerf} to validate the
effectiveness of adjusting sampling point locations using the proposed
method.
%
%
We used one-eighth of the images from each scene as test images~\cite{mildenhall2021nerf}.
%
The training configuration for the proposed method's network involved
using $1,008$ rays, i.e., $N_r=1,008$, chosen randomly from 
$1,008 \times 756$ rays.
Each ray had 128 sampling points, i.e., $N_s=128$.
%
%
The sampling module $G$ implemented an MLP
structure with three layers of the sampling blocks outlined in
Fig.~\ref{fig:sampling}.
The hidden layer of the ray-wise MLP $G^{(ray)}$ contains
1,024 units, while the scene-wise MLP $G^{(scene)}$ has 4,032
units in its hidden layer.
Additionally, a Sigmoid function was incorporated into the output
layer of $G$.

This experiment employed Peak Signal to Noise Ratio (PSNR),
Structural SIMilarity (SSIM), and Learned Perceptual Image Patch
Similarity (LPIPS)~\cite{zhang2018unreasonable} as the benchmarks for
assessing rendered image quality.
%
%


Table~\ref{table:result} shows the results for the eight scenes.
Results are highlighted in bold, where significant differences were
observed through the t-test with the significance level of 5\%.
Our method surpassed conventional NeRF 
%
in terms of PSNR, SSIM, and LPIPS
in the Room, Fern, Leaves, and T-rex scenes.

Fig.~\ref{fig:result2} (a) to (c) demonstrate rendering examples from the test
view.
%
The results
revealed that our method succeeded in rendering leaves that were not
drawn by conventional NeRF.

The distribution of sampling points shown in Fig.~\ref{fig:result2}(d)
illustrates that
%
sampling points tend to gather in areas
thought to contain objects, contributing to the rendering of images
with reduced artifacts.

\section{Conclusions}

This study introduces a learning-based method for optimizing sampling
points in novel viewpoint image generation using NeRF.
%
%
Our method adjusts the placement of sampling points based on scene
properties, potentially reducing artifacts and enhancing
image quality compared with conventional NeRF.
Experiments with real images demonstrated the effectiveness of the
proposed method in scene representation.

\bibliographystyle{unsrt}
\bibliography{reference}


\end{document}